\documentclass[lettersize,journal]{IEEEtran}
\usepackage{amsmath,amsfonts}
\usepackage{algorithmic}
\usepackage{algorithm}
\usepackage{array}
\usepackage[caption=false,font=normalsize,labelfont=sf,textfont=sf]{subfig}
\usepackage{textcomp}
\usepackage{stfloats}
\usepackage{url}
\usepackage{verbatim}
\usepackage{graphicx}
\usepackage{cite}
\usepackage{subcaption}
\usepackage{pifont}
\usepackage{multirow}
\usepackage{booktabs} 
\usepackage{adjustbox}

\begin{document}

\title{MR-UBi: Mixed Reality-Based Underwater Robot Arm Teleoperation System with Reaction Torque Indicator via Bilateral Control}
\author{
Kohei~Nishi$^{\dag 1}$, Masato Kobayashi$^{\dag 1,2*}$, Yuki Uranishi$^{1}$
\thanks{
$^{\dag}$ Equal Contribution, 
$^{1}$ The University of Osaka, 
$^{2}$ Kobe University,\\
$^*$ corresponding author: kobayashi.masato.cmc@osaka-u.ac.jp
}
}

\maketitle
\begin{abstract}
We present a mixed reality-based underwater robot arm teleoperation system with a reaction torque indicator via bilateral control (MR-UBi). 
The reaction torque indicator (RTI) overlays a color and length-coded torque bar in the MR-HMD, enabling seamless integration of visual and haptic feedback during underwater robot arm teleoperation. 
User studies with sixteen participants compared MR-UBi against a bilateral-control baseline. 
MR-UBi significantly improved grasping-torque control accuracy, increasing the time within the optimal torque range and reducing both low and high grasping torque range during lift and pick-and-place tasks with objects of different stiffness.
Subjective evaluations further showed higher usability (SUS) and lower workload (NASA--TLX).
Overall, the results confirm that \textit{MR-UBi} enables more stable, accurate, and user-friendly underwater robot-arm teleoperation through the integration of visual and haptic feedback.
For additional material, please check: \url{https://mertcookimg.github.io/mr-ubi}
\end{abstract}

\section{Introduction}
\IEEEPARstart{U}{nderwater} operations such as marine resource development, deep-sea exploration, and infrastructure maintenance have gained increasing importance in recent years~\cite{tang2024rov6, kang2020development, pi2021twinbot}.  
Due to high pressure, low temperature, and poor visibility, direct human intervention remains risky and costly, creating a strong demand for underwater teleoperation systems~\cite{liu2024selfimprovingautonomousunderwatermanipulation, sun2024uwr}.  

Various robot arm teleoperation systems have been proposed, including keyboard or joystick control~\cite{ozdamar2022shared, qi2025development}, VR controllers~\cite{yim2022wfh}, and motion-capture systems~\cite{kang2025catch}.  
Among them, the leader–follower teleoperation system offers intuitive operation by reproducing the operator’s motion on a remote robot arm and has been widely adopted in industrial and medical applications~\cite{diez2018evaluation}.  
Leader–follower systems are typically divided into \textit{unilateral} and \textit{bilateral} control architectures.  
Unilateral control transmits only position commands, forcing operators to infer contact forces from visual cues, which increases cognitive load and reduces precision~\cite{zhao2023learningfinegrainedbimanualmanipulation}.
In contrast, bilateral control exchanges both position and force, allowing the operator to feel reaction forces/torque from the environment in real time~\cite{mk2025alpha}.
Although bilateral control theoretically provides haptic feedback in underwater environments, the perception of reaction torque becomes ambiguous due to hydrodynamic disturbances and limited training time~\cite{hagen2024beyond}.  
Therefore, enhancing sensory feedback, particularly through visual cues, has become an important research direction.
\begin{figure}[t]
    \centering
    \includegraphics[width=0.9\linewidth]{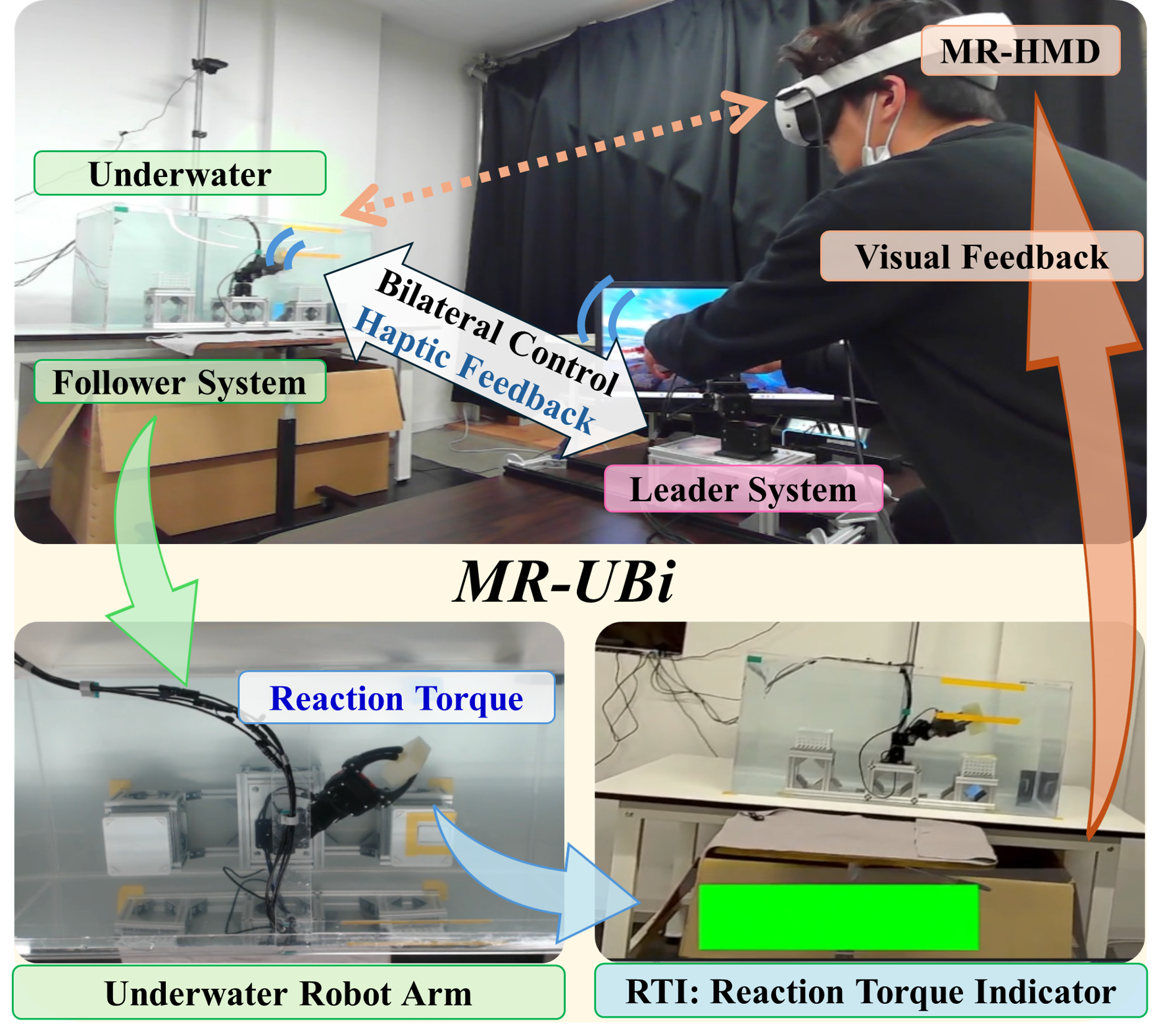}
    \caption{Overview of Proposed MR-UBi}
    \label{fig:teaser}
\end{figure}

Recent advances in mixed reality (MR) show that overlaying task-relevant information directly in the operator’s view can enhance situational awareness and reduce cognitive effort~\cite{van2024augmenting, felten2025advancing}. In land teleoperation, haptic-based visual feedback techniques have been widely explored to augment or substitute haptic sensations under unilateral control.
Although bilateral systems for underwater robotic arms exist, most previous implementations employ 1 degrees of freedom (DoF) robot arm~\cite{motoi2023twobiuwr,motoi2018development} that prioritize simplicity and stability, while high-end systems such as Ocean One~\cite{oceanone2016khatib,oceanone2020brantner} achieve precise force rendering at the cost of complexity and expense. 
To the best of our knowledge, intuitive haptic-based visual feedback remains largely unexplored in underwater teleoperation, even though such cues could compensate for ambiguous haptic sensations and help non-expert operators regulate grasping force more effectively.

Therefore, this paper presents MR-UBi, a mixed reality-based underwater robotic arm teleoperation system with a reaction torque indicator via bilateral control, as shown in Fig.~\ref{fig:teaser}.
Our system introduces a reaction torque indicator (RTI) that encodes torque magnitude and status using color and bar-length visualization within the MR-HMD view, complementing haptic feedback and enabling intuitive force regulation. MR-UBi comprises three modules: (1) a low-cost 3-DoF underwater robotic arm, (2) leader–follower bilateral control mechanism that provides haptic feedback, and (3) RTI module that enables haptic-based visual feedback via MR-HMD passthrough.

The main contributions are summarized as follows:
\begin{itemize}
    \item MR-UBi is a mixed reality–based underwater teleoperation system that integrates 3-DoF underwater robot arms, four-channel bilateral control, and MR-HMD passthrough with RTI.
    \item RTI provides intuitive in-view haptic-based visual feedback by encoding grasping torque through bar length and a hybrid continuous–discrete hue scheme.
    \item Experimental results show that MR-UBi improves torque control accuracy and stability while enhancing usability and reducing workload.
\end{itemize}

\nocite{zhao2023learningfinegrainedbimanualmanipulation,mk2025alpha,brown2025rov,sitler2024uvms,chen2025subsensevrhapticmotorfeedback,ozdamar2022shared,TIAN2025BI,motoi2018development,motoi2023twobiuwr,oceanone2016khatib,oceanone2020brantner,moortgat2022rift,mielke2025sensary}

\begin{table*}[t]
\centering
\caption{Comparison of Teleoperation Interfaces Across Land and Underwater Environments}
\label{tab:ui_diff}
\small
\renewcommand{\arraystretch}{1.05}
\setlength{\tabcolsep}{4pt} 
\begin{adjustbox}{width=\linewidth}
\begin{tabular}{lccccccccc}
\hline
\textbf{System} & \textbf{Env.} & \textbf{Operator Interface} & \textbf{Control} &
\textbf{Haptic Feedback} & \multicolumn{4}{c}{\textbf{Haptic-Based Visual Feedback}} \\
\cline{6-9}
 &  &  &  &  & \textbf{Cue} & \textbf{Encoding} & \textbf{Display} & \textbf{Overlay} \\
\hline
ALOHA~\cite{zhao2023learningfinegrainedbimanualmanipulation} & Land & Leader (6-DoF + Gripper) & Uni. & -- & -- & -- & -- & -- \\
ALPHA-$\alpha$~\cite{mk2025alpha} & Land & Leader (6-DoF + Gripper) & Bi. & Torque & -- & -- & -- & -- \\
A-RIFT~\cite{moortgat2022rift} & Land & Keyboard + Mouse & Uni. & -- & Bar & Continuous (Length \& Hue) & 2D Screen & \ding{51} \\
SensARy-Bar~\cite{mielke2025sensary} & Land & Hand Tracking & Uni. & Vibration & Bar & Continuous (Length) \& Discrete (3 RGB) & HMD (PT) & \ding{51} \\
SensARy-Arrow~\cite{mielke2025sensary} & Land & Hand Tracking & Uni. & Vibration & Arrow & Continuous (Direction \& Hue) & HMD (PT) & \ding{51} \\
UVMS~\cite{sitler2024uvms} & Underwater & Leader (3D Systems Touch) & Uni. & -- & -- & -- & -- & -- \\
SubSense~\cite{chen2025subsensevrhapticmotorfeedback} & Underwater & VR Controller + Glove & Uni. & Vibration & -- & -- & -- & -- \\
Module~\cite{motoi2023twobiuwr,motoi2018development} & Underwater & Leader (1-DoF) & Bi. & Torque & -- & -- & -- & -- \\
Ocean One~\cite{oceanone2016khatib,oceanone2020brantner} & Underwater & Leader (Sigma-7) & Bi. & Force/Torque & Numeric & Continuous (Graph) & 2D Screen & \ding{55} \\
\textbf{MR-UBi (Ours)} & \textbf{Underwater} & \textbf{Leader (3-DoF + Gripper)} & \textbf{Bi.} & \textbf{Torque} &
\textbf{Bar} & \textbf{Continuous (Length) \& Hybrid (Hue)} & \textbf{HMD (PT)} & \textbf{\ding{51}} \\
\hline
\end{tabular}
\end{adjustbox}
\vspace{3pt}
\scriptsize
\noindent
\textbf{Abbreviations:}  
Env.:Environment,
Uni.: Unilateral control, Bi.: Bilateral control,  
HMD (PT): Head-Mounted Display (Passthrough),  
DoF: Degrees of Freedom.
\end{table*}

\section{Related Work}
\subsection{Teleoperation in Land and Underwater Environments}
Teleoperation systems are generally categorized into \textit{unilateral} and \textit{bilateral} control architectures.
A unilateral system transmits position commands from the leader to the follower, such as ALOHA~\cite{zhao2023learningfinegrainedbimanualmanipulation}, offering implementation simplicity and stability.
However, because no force feedback is returned, the operator must infer contact conditions solely from visual information, often resulting in excessive grasping force, object damage, or reduced precision.
In contrast, bilateral control exchanges both position and force, allowing the operator to feel environmental reactions in real time and regulate applied forces more accurately, such as ALPHA-$\alpha$~\cite{mk2025alpha}.

Underwater teleoperation introduces additional challenges such as hydrodynamic drag, buoyancy, limited visibility, and sealing constraints, all of which degrade force sensing and haptic fidelity~\cite{brown2025rov}.
For this reason, many underwater systems adopt unilateral control with camera monitoring (e.g., UVMS~\cite{sitler2024uvms}, SubSense~\cite{chen2025subsensevrhapticmotorfeedback}) to prioritize stability under uncertain forces.
Although bilateral control has been explored to convey resistance and stiffness~\cite{TIAN2025BI,motoi2018development,motoi2023twobiuwr}, most implementations employ low-DoF prototypes (typically 1-DoF) focused on mechanical simplicity and robustness.
At the other extreme, high-end systems such as \textit{Ocean One}~\cite{oceanone2016khatib,oceanone2020brantner} achieve highly accurate force rendering using 7-DoF arms with force/torque sensors and Sigma-7 devices, but at the cost of considerable system complexity and expense.

The proposed MR-UBi system situates itself between these extremes, achieving four-channel bilateral control on a compact 3-DoF arm with a gripper via sensorless torque estimation at 1 kHz.
Each arm can be fabricated for approximately USD 3{,}400, providing a cost-efficient yet dynamically stable bilateral teleoperation platform suitable for underwater operation.
The following section reviews how such systems have incorporated haptic-based visual feedback mechanisms to support haptic perception.

\subsection{Haptic-Based Visual Feedback for Teleoperation}
When direct haptic perception is limited, haptic-based visual feedback conveys interaction forces through graphical or vibrotactile cues.
A-RIFT~\cite{moortgat2022rift} presented such cues on a 2D screen, using a keyboard-and-mouse interface.
The normalized force was mapped to a color bar whose length and hue changed continuously with the applied force.
Although effective for coarse awareness, such displays required gaze shifts and served as substitutes rather than complements to haptics.

Recent studies have shifted toward in-view visual augmentation using HMDs, overlaying feedback cues directly within the operator’s field of view~\cite{van2024augmenting,felten2025advancing}.  
SensARy~\cite{mielke2025sensary} introduced two variants:  
(i) SensARy-Bar, encoding force magnitude through discrete color levels, and  
(ii) SensARy-Arrow, showing arrow length and direction with continuous hue gradients.  
Displayed in an HMD passthrough view and supported by vibrotactile cues, SensARy improved situational awareness without gaze shifts.  
Among them, \textit{SensARy-Bar} achieved the best performance~\cite{mielke2025sensary}, offering clear discrete color transitions that intuitively conveyed force states (low, optimal, high).  
However, its fully discrete encoding hindered the perception of gradual force variation, limiting fine modulation.  
Moreover, since SensARy operated under unilateral control with vibrotactile feedback only, no actual force reflection was provided, restricting its effectiveness in precise manipulation tasks.
A comparison of representative teleoperation interfaces across land and underwater domains is summarized in Table~\ref{tab:ui_diff}.

In contrast, underwater systems employ bilateral control but lack haptic-based visual feedback.  
As summarized in Table~\ref{tab:ui_diff}, Module Arm~\cite{motoi2018development,motoi2023twobiuwr} implements bilateral control without visual overlays.  
Advanced platforms such as \textit{Ocean One}~\cite{oceanone2016khatib,oceanone2020brantner} provide multi-screen visualization, where stereoscopic camera images are shown on a 3D display for immersive perception, while the robot’s state and force/torque information are rendered on separate 2D panels.  
This spatial separation between haptic and visual channels often induces gaze shifts and disrupts smooth force regulation.

MR-UBi bridges these gaps by combining (1) a low-cost 3-DoF underwater robotic arm, (2) bilateral control that provides haptic feedback, and (3) RTI module that enables haptic-based visual feedback via MR-HMD.
Especially, RTI encodes torque magnitude (bar length) and state (color) using a hybrid continuous and discrete hue scheme.  
Intermediate colors are linearly interpolated between RGB states, allowing users to perceive torque changes intuitively.  

\begin{figure*}[t]
\centering
\includegraphics[width=0.95\linewidth]{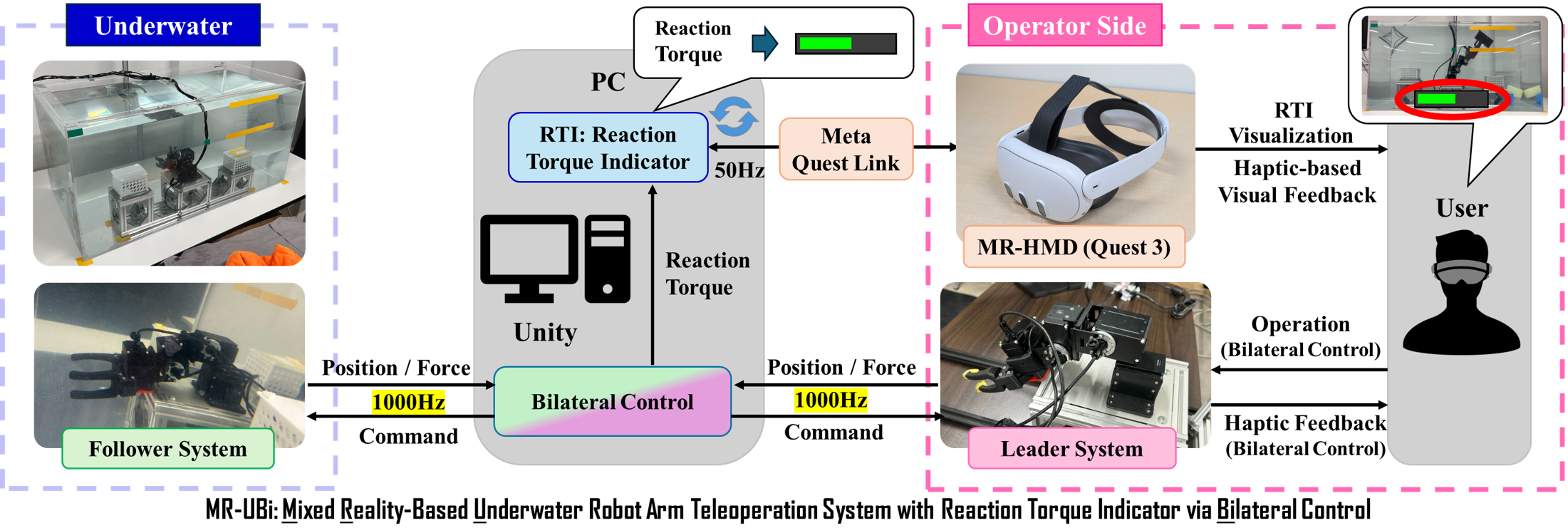}
\caption{Overview of MR-UBi, Left: Follower System (Underwater), Right: Leader System (User), Center: Unity-based System.}
\label{fig:system_diagram}
\end{figure*}
\begin{figure}[t]
    \centering
    \includegraphics[width=0.9\linewidth]{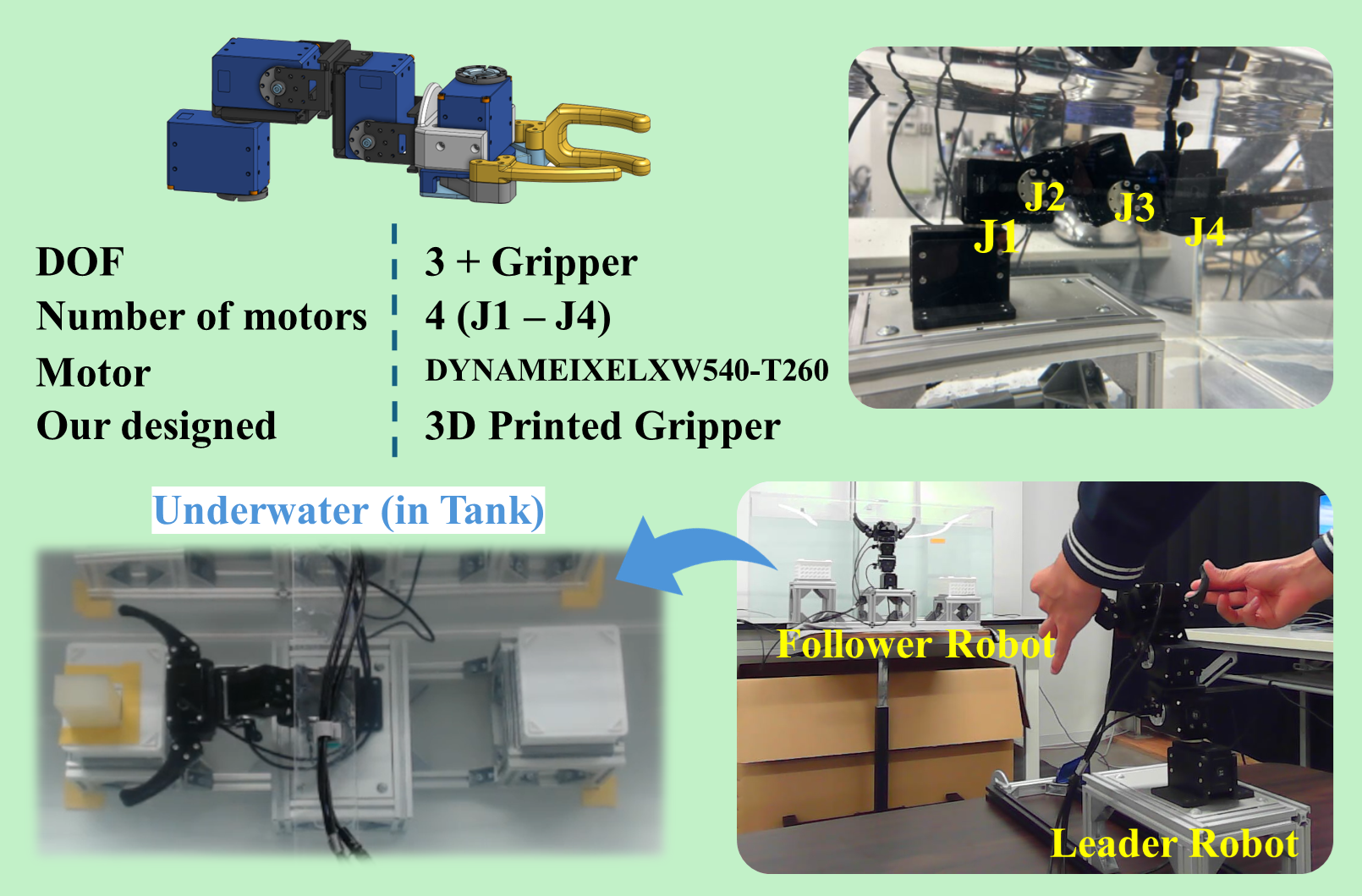}
    \caption{Underwater Robot Arm (MR-UBi Arm)}
    \label{fig:underwater-robot-arm-sys}
\end{figure}

\section{MR-UBi: Mixed Reality-Based Underwater Robot Arm Teleoperation System with Reaction Torque Indicator via Bilateral Control}
\subsection{Overview}
This paper proposes MR-UBi, a mixed reality-based underwater robot arm teleoperation system with a reaction torque indicator via bilateral control.
MR-UBi integrates three primary components: (1) a low-cost 3-DoF underwater robotic arm, (2) leader–follower bilateral control mechanism that provides haptic feedback, and (3) RTI module that enables haptic-based visual feedback via MR-HMD passthrough.
The overall system configuration is shown in Fig.~\ref{fig:system_diagram}. 

During operation, the operator observes the real-world workspace and the underwater robotic arm through the MR-HMD’s video passthrough. 
Simultaneously, the RTI is overlaid at the bottom-center of the field of view, providing real-time visual feedback of grasping torque. 
Leader and follower robot arms consist of 3-DoF and a gripper, each actuated by a waterproof Dynamixel XW540-T260 motor. 
The leader–follower synchronization is achieved through 4-channel bilateral control algorithm. 
Reaction torque data are continuously estimated and reflected in the RTI, enabling the operator to perceive the underwater interaction intuitively.

\subsection{Underwater Robot Arm}
The developed leader–follower underwater robotic arm employs fully waterproof motors (Dynamixel XW540-T260, IP68 standard) for each joint, as shown in  Fig.~\ref{fig:underwater-robot-arm-sys}.
The mechanism comprises three rotational joints and a two-fingered gripper (end-effector), inspired by the structure in~\cite{yamamoto2024gripper}. 
The links and frames are fabricated using aluminum profiles and 3D-printed components for lightweight durability. 
The per-arm parts cost is approximately USD 3,400.

\begin{figure}[t]
    \centering
    \includegraphics[width=\linewidth]{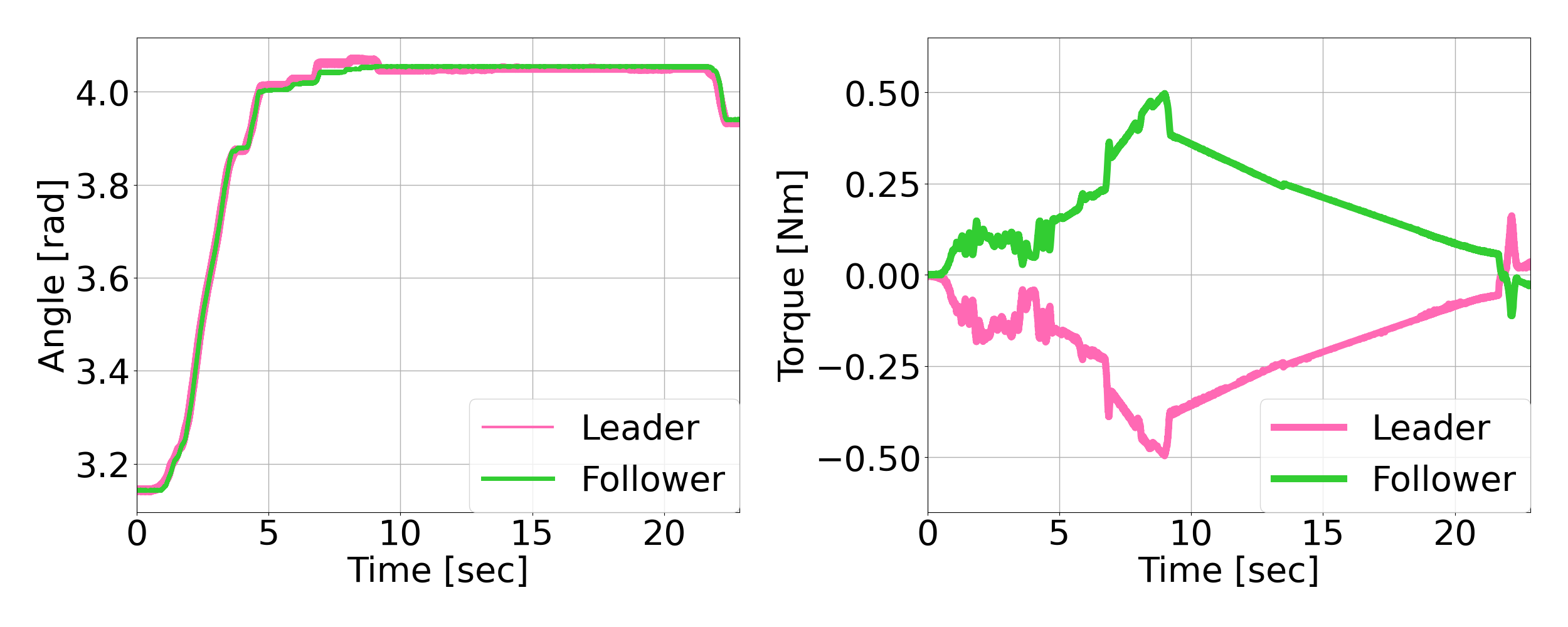}
    \caption{Example of Bilateral Control Data (J4)}
    \label{fig:underwater-bi-test}
\end{figure}
\begin{figure*}[t]
\centering
\includegraphics[width=\linewidth]{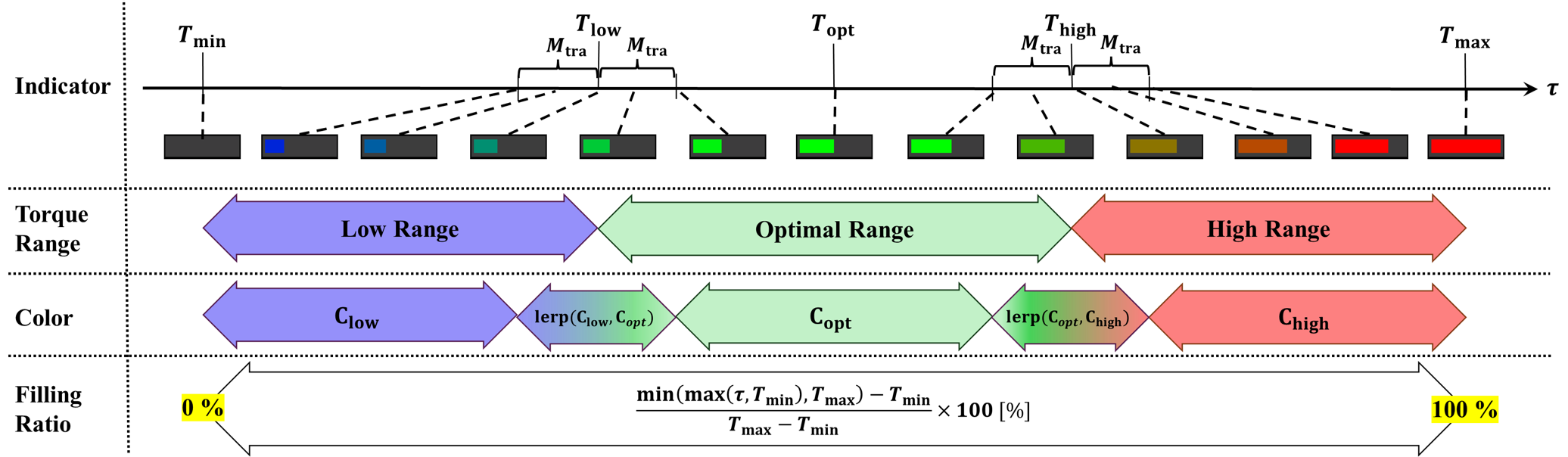}
\caption{Design of Proposed Reaction Torque Indicator (RTI)}
\label{fig:indicator_design}
\end{figure*}
\begin{figure*}[t]
\centering
\includegraphics[width=\linewidth]{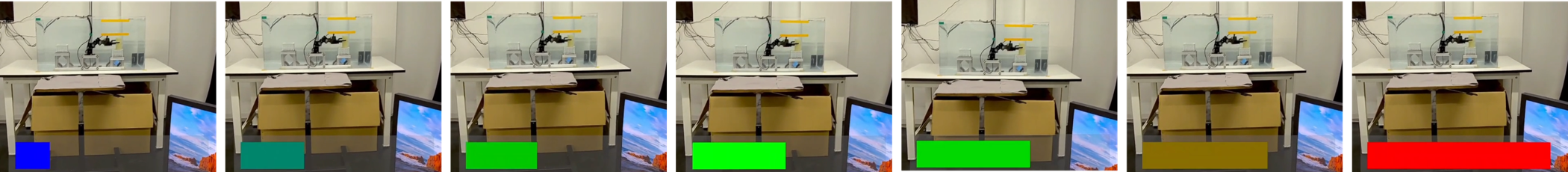}
\caption{Example of RTI via MR-HMD Video Passthrough}
\label{fig:indicator_example}
\end{figure*}

\subsection{Bilateral Control}
A four-channel bilateral control is achieved through position tracking and the use of action-reaction principles as follows.
\begin{align}
\theta_l - \theta_f &= 0,\\
\tau_l + \tau_f &= 0,
\end{align}
where $\theta$ and $\tau$ denote the joint angles and torques of the leader $\bigcirc_l$ and follower $\bigcirc_f$, respectively. The torque value is estimated in a sensorless method using a Reaction Torque Observer (RTOB)~\cite{RTOB}. 

In MR-UBi, four-channel bilateral control is applied to all four joints and operates at $1,\mathrm{kHz}$, ensuring fast and stable underwater performance.
As an example, Fig.~\ref{fig:underwater-bi-test} shows that the leader and follower grippers maintain accurate position tracking and satisfy the action–reaction relationship.

\subsection{Reaction Torque Indicator Design}
Although bilateral control provides real-time haptic feedback, subtle torque variations are often difficult to perceive underwater, particularly for inexperienced operators.  
The proposed RTI visually supplements the reflected torque within the MR-HMD, enabling intuitive torque awareness and stable manipulation.

The RTI represents reaction torque using a \textit{dual-representation} scheme that combines two types of visual cues:
(1) the \textit{bar length}, which shows the torque magnitude quantitatively, and
(2) the \textit{color hue}, which indicates whether the torque is within an appropriate range.
Unlike previous discrete color-based designs~\cite{mielke2025sensary}, MR-UBi employs a continuous bar length and a hybrid continuous–discrete hue transition to represent torque magnitude and state more intuitively.
This approach allows operators to perceive gradual torque changes smoothly, while still clearly distinguishing optimal range.

The bar filling ratio $L(\tau)$ is calculated by normalizing the reaction torque $\tau$ within the range between its minimum $T_{\text{min}}$ and maximum $T_{\text{max}}$ values, as defined in Eq.~(\ref{eq:Filling}):
\begin{equation}
L(\tau) = \frac{\min(\max(\tau, T_{\text{min}}), T_{\text{max}}) - T_{\text{min}}}{T_{\text{max}} - T_{\text{min}}} \times 100
\label{eq:Filling}
\end{equation}
This normalization enables the operator to intuitively recognize the relative torque magnitude at a glance.

To explicitly indicate the appropriateness of the applied torque, color transitions are employed. 
The torque range is divided into three range: \textit{Low} ($\tau \leq T_{\mathrm{low}}$), \textit{optimal} ($T_{\mathrm{low}} \leq \tau \leq T_{\mathrm{high}}$), and \textit{High} ($T_{\mathrm{high}} \leq \tau$). 
Each range is associated with a representative color: $\mathbf{C}_{\mathrm{low}}$, $\mathbf{C}_{\mathrm{opt}}$, and $\mathbf{C}_{\mathrm{high}}$, respectively. 
A transition margin $M_{\text{tot}}$ is introduced to achieve smooth color blending between neighboring range, as defined by Eq.~(\ref{eq:margin}):
\begin{align}
M_{\text{tot}} &= 2 M_{\mathrm{tra}}
\label{eq:margin}
\end{align}
The final color $\mathbf{C}(\tau)$ is computed using a linear interpolation function $\operatorname{lerp}(\mathbf{a}, \mathbf{b}; \alpha) = (1 - \alpha)\mathbf{a} + \alpha \mathbf{b}$, as follows:
\begin{equation}
\resizebox{\columnwidth}{!}{$
\mathbf{C}(\tau)=
\begin{cases}
\mathbf{C}_{\mathrm{low}}, &
  \tau \le T_{\mathrm{low}}\!-\!M_{\mathrm{tra}},\\
\operatorname{lerp}\!\bigl(\mathbf{C}_{\mathrm{low}},\mathbf{C}_{\mathrm{opt}};\alpha_1(\tau)\bigr), &
  T_{\mathrm{low}}\!-\!M_{\mathrm{tra}}<\tau<T_{\mathrm{low}}\!+\!M_{\mathrm{tra}},\\
\mathbf{C}_{\mathrm{opt}}, &
  T_{\mathrm{low}}\!+\!M_{\mathrm{tra}}\le\tau\le T_{\mathrm{high}}\!-\!M_{\mathrm{tra}},\\
\operatorname{lerp}\!\bigl(\mathbf{C}_{\mathrm{opt}},\mathbf{C}_{\mathrm{high}};\alpha_2(\tau)\bigr), &
  T_{\mathrm{high}}\!-\!M_{\mathrm{tra}}<\tau<T_{\mathrm{high}}\!+\!M_{\mathrm{tra}},\\
\mathbf{C}_{\mathrm{high}}, &
  T_{\mathrm{high}}\!+\!M_{\mathrm{tra}}\le\tau
\end{cases}
$}
\label{eq:Color}
\end{equation}
\begin{equation}
\alpha_i(\tau)=M_{\mathrm{tot}}^{-1}\!\bigl(\tau-(T_i-M_{\mathrm{tra}})\bigr),\quad i\in\{1,2\}.
\label{eq:alpha}
\end{equation}

Through this design, operators can simultaneously perceive the \textit{magnitude} of the applied torque from the bar length and the quality of torque control (low, optimal, or high) from the color hue. 
The introduction of a transition margin ensures smooth visual transitions near the threshold range, preventing abrupt color changes and allowing the operator to finely modulate grasping torque.

As shown in Fig.~\ref{fig:indicator_example}, the RTI is superimposed on the MR-HMD display to visualize the real-time reaction torque.  
This visualization enables safe and precise manipulation by maintaining optimal grasping torque, reducing the risk of object damage, and improving operational accuracy.

\begin{figure*}[t]
\centering
\includegraphics[width=\linewidth]{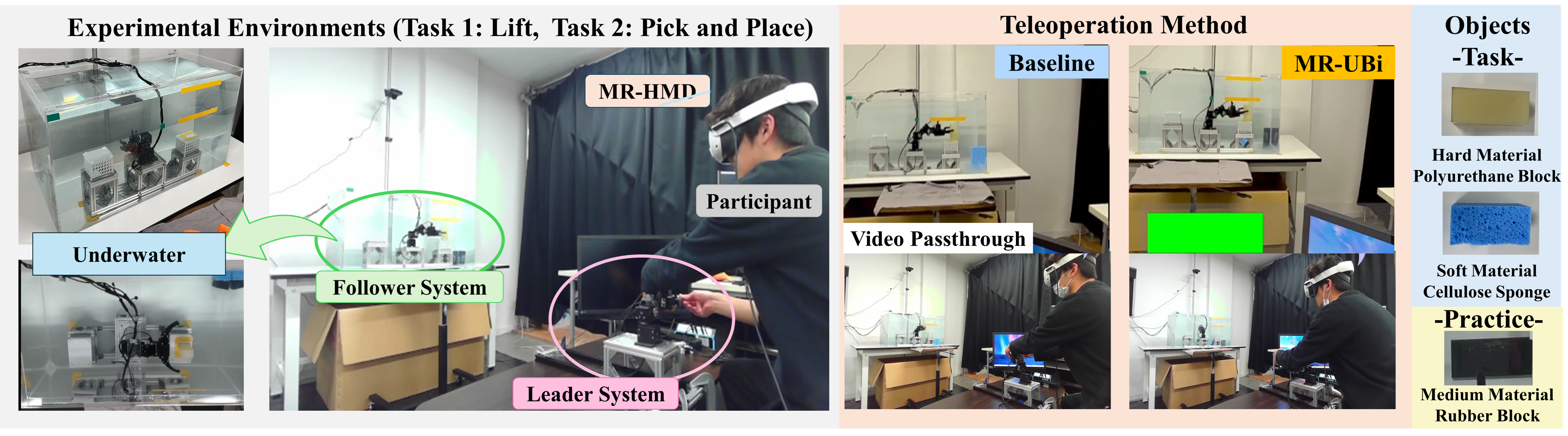}
\caption{Experimental Environments and Conditions}
\label{fig:env}
\end{figure*}

\section{Experiments}
This section evaluates the overall effectiveness of the proposed MR-UBi, focusing on how the Reaction Torque Indicator (RTI) contributes to torque regulation and user experience in underwater teleoperation. 
Specifically, we examine how the RTI visually complements the haptic feedback provided by bilateral control, thereby enabling operators to perceive reaction torque more intuitively and to regulate grasping torque with higher precision. 
We compare two configurations: a bilateral-control teleoperation system with MR-HMD video passthrough as the \textit{Baseline} (conventional method) and the MR-UBi (proposed method), as illustrated in Fig.~\ref{fig:env} and Table~\ref{tab:mrubi_layers}. 

The evaluation addresses the following objectives:
\begin{itemize}
  \item (1) MR-UBi improves grasping-torque control accuracy by increasing the time within the optimal torque range and reducing torque error.  
  \item (2) MR-UBi enhances perceived usability and lowers operator workload compared with the baseline.
\end{itemize}

\subsection{Experimental Design}
\begin{table}[t]
\centering
\caption{System Component of Baseline and MR-UBi}
\label{tab:mrubi_layers}
\begin{tabular}{l|c|c}
\hline
\textbf{Component} & \textbf{Baseline} & \textbf{MR-UBi} \\
\hline
Bilateral Control & \ding{51} & \ding{51} \\
HMD Video Passthrough & \ding{51} & \ding{51} \\
RTI Visual Overlay & -- & \ding{51} \\
\hline
\end{tabular}
\end{table}

\begin{figure}[t]
    \centering
    \includegraphics[width=\linewidth]{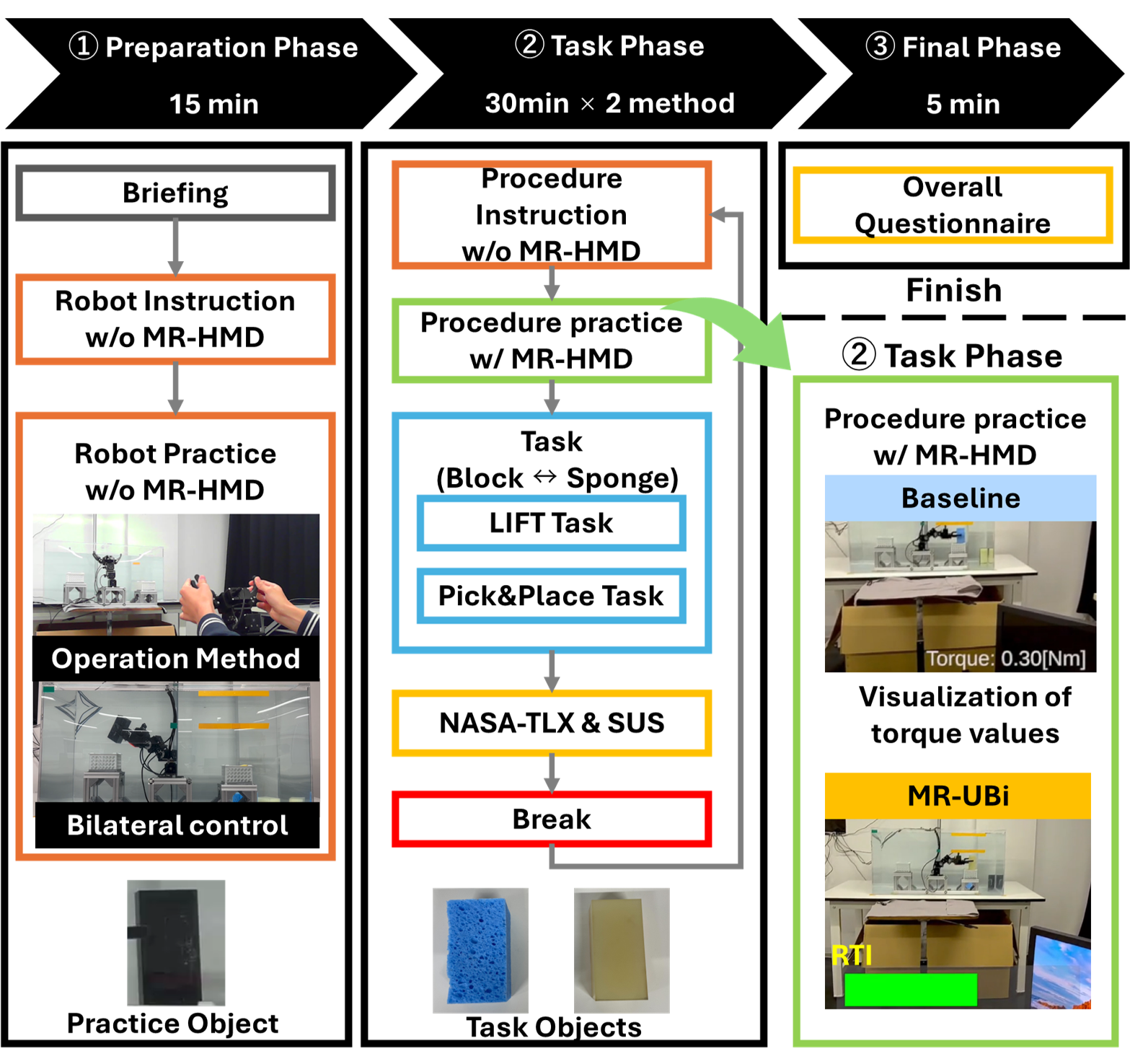}
    \caption{Experimental Flow}
    \label{fig:ex-flow}
\end{figure}

To evaluate the overall effectiveness of MR-UBi, we conducted a within-subject experiment consisting of three sequential phases: \textit{Preparation}, \textit{Task}, and \textit{Final} (Fig.~\ref{fig:ex-flow}).  
The independent variable was the system configuration {\it Baseline} and {\it MR-UBi}.  
Dependent variables were (i) manipulation accuracy and (ii) subjective ratings of usability and workload.

\paragraph{Preparation Phase (15\,min).}
Participants first received a briefing on the study procedure and safety, followed by robot operation practice using bilateral control without the MR-HMD.  
To avoid task-specific habituation, a training block made of rubber with a different stiffness from the main experimental objects was used during this practice.  
This ensured that all participants could operate the leader–follower system safely before being exposed to the MR-HMD passthrough environment.
\begin{figure}[t]
    \centering
    \includegraphics[width=\linewidth]{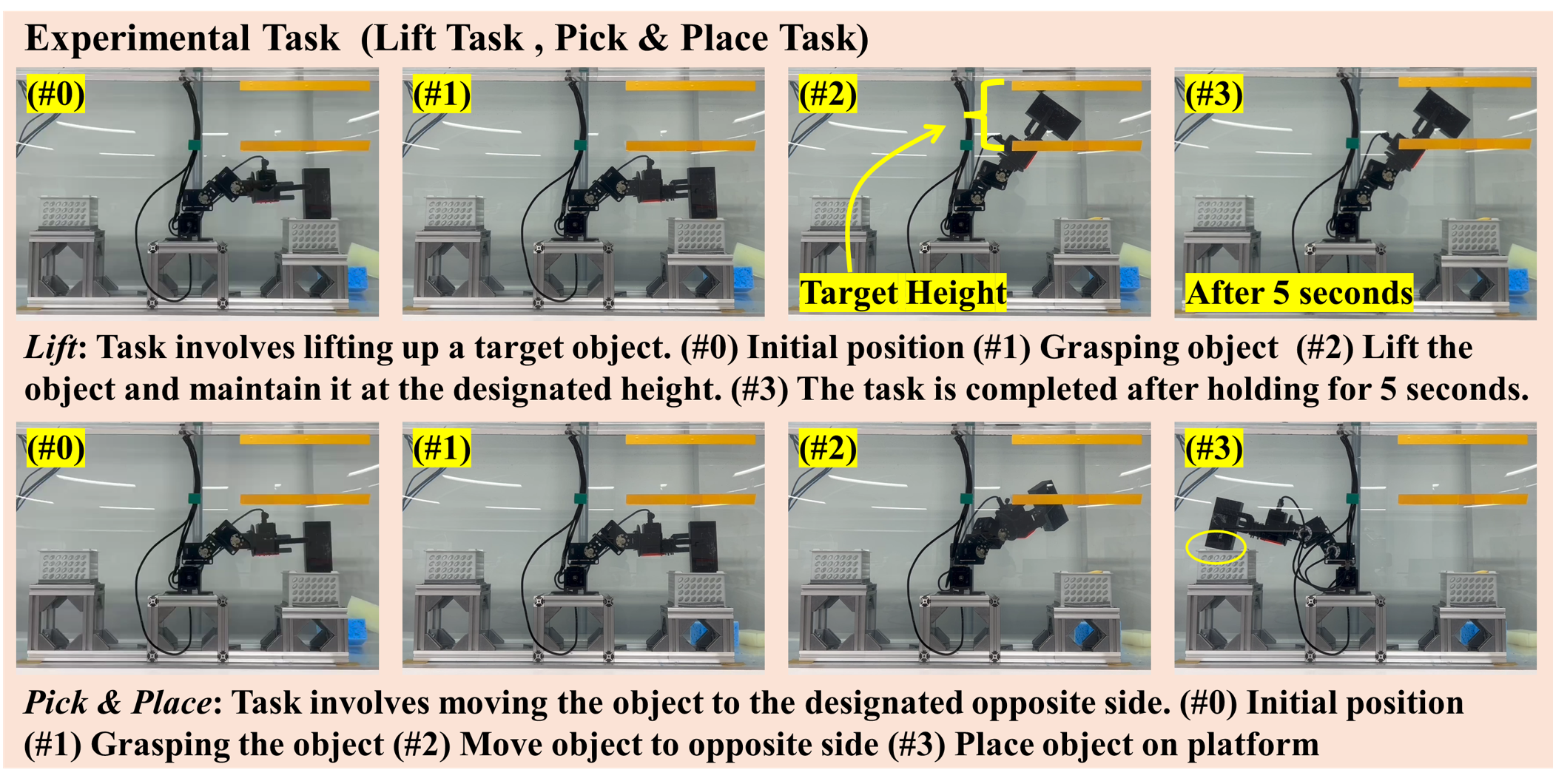}
    \caption{Experimental Task Flow}
    \label{fig:taskflow}
\end{figure}

\paragraph{Task Phase (30\,min × 2 method).}
Participants performed all conditions in counterbalanced order.
Each condition began with a procedure practice using the MR-HMD, after which participants performed two manipulation tasks,{\it Lift} and {\it Pick\&Place}, as shown in Fig.~\ref{fig:taskflow}.
Two target objects were used in the formal trials: a rigid polyurethane block and a compliant cellulose sponge, representing different stiffness levels.  
Short breaks were inserted between methods to minimize fatigue.  
In the Baseline condition, a numeric torque display was presented only during the initial calibration phase and was hidden throughout the actual trials.
This calibration feedback was provided so that participants could familiarize themselves with the optimal range of reaction torque, as the Baseline configuration did not include the RTI and therefore offered no visual torque cues during task execution.
In the MR-UBi condition, the RTI was continuously overlaid in the MR-HMD view, visualizing grasping torque through bar length and color cues.
After completing each condition, participants immediately filled out the System Usability Scale (SUS)~\cite{sus} and NASA Task Load Index (NASA--TLX)~\cite{nasa-tlx} questionnaires to assess usability and workload for that specific condition.

\paragraph{Final Phase (5\,min).}
After both methods, participants completed an overall questionnaire and provided free-text comments.  

\subsection{Experimental Environment}
The experimental setup comprised a water tank and a remotely operated leader robot system placed within the same laboratory environment, as shown in Fig.~\ref{fig:env}.
An underwater follower robot was rigidly mounted on a pedestal fixed to the bottom of the tank.
The leader robot system was installed on a desk approximately 3 m away from the water tank, where the operator controlled the system while wearing the MR-HMD (Quest~3).

The leader-follower system via bilateral control runs at $1\,\mathrm{kHz}$; Unity handles communication, visualization, and MR rendering. 
The RTI is persistently displayed as a bar at the bottom of the MR-HMD view whose length and color vary with the estimated torque. 
RTI parameters are defined as follows.
$T_{\min}=0.00\,\mathrm{Nm}$, $T_{\max}=0.60\,\mathrm{Nm}$, $T_{\mathrm{opt}}=0.30\,\mathrm{Nm}$; optimal band $[T_{\mathrm{low}},T_{\mathrm{high}}]=[0.20,0.40]\,\mathrm{Nm}$ with transition margin $M_{\mathrm{tra}}=0.05\,\mathrm{Nm}$; colors:
$\mathbf{C}_{\mathrm{low}}=(0,0,255)$,
$\mathbf{C}_{\mathrm{opt}}=(0,255,0)$,
$\mathbf{C}_{\mathrm{high}}=(255,0,0)$.
\subsection{Participants}
Sixteen participants (3 female, 13 male) took part in the study.  
Their MR-HMD experience levels were distributed as follows: 4 had no prior experience, 11 had used it a few times, and 1 used it frequently.
Only one participant had prior experience operating another robotic arm, while all others were novices.  
Before the experiment, participants received detailed instructions on the operation and safety procedures and provided written informed consent in accordance with institutional guidelines.
\begin{figure}[t]
    \centering
    \includegraphics[width=0.9\linewidth]{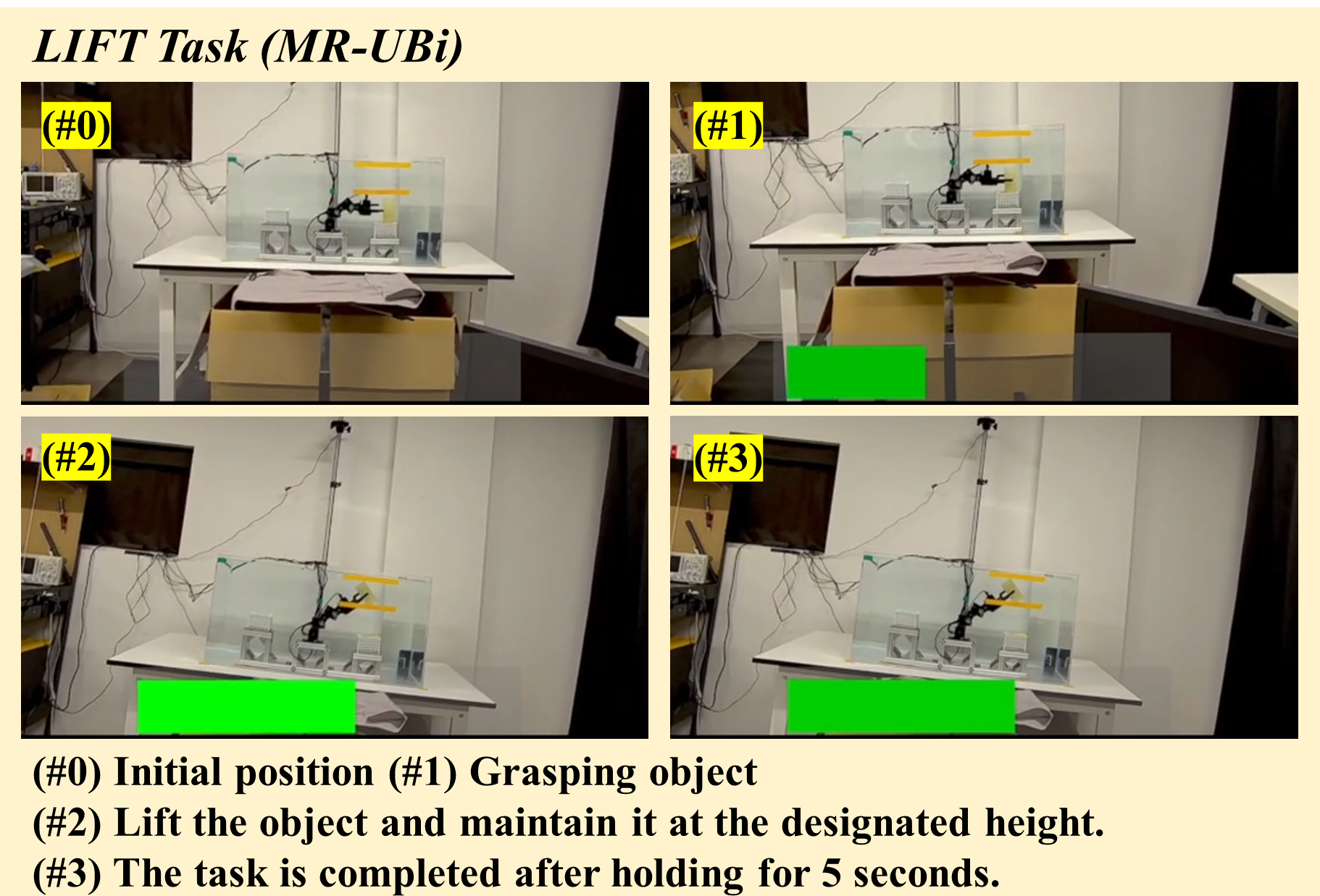}
    \caption{Lift Task using MR-UBi}
    \label{fig:LIFT_with_indica}
\end{figure}

\subsection{Evaluation Metrics}
\subsubsection{Quantitative Metrics}

To evaluate the precision of grasping-torque regulation, four quantitative indices were computed from the grasping-torque time series of each trial:  
\begin{itemize}
  \item Optimal [\%]: Proportion of samples within the optimal torque range $[T_{\mathrm{low}}, T_{\mathrm{high}}]=[0.20, 0.40]\,\mathrm{Nm}$.
  \item Low [\%]: Proportion of samples in which the grasping torque was below the lower bound of the optimal range.
  \item High [\%]: Proportion of samples in which the grasping torque exceeded the upper bound of the optimal range.
  \item MAE [Nm]: Mean absolute error from the optimal torque $T_{\mathrm{opt}}=0.30\,\mathrm{Nm}$.
\end{itemize}

Here, \textit{Low [\%]} quantifies the fraction of samples below the optimal range, corresponding to the risk of slippage or unstable grasping.  
\textit{Optimal [\%]} indicates the time proportion during which the applied torque remained within the desired range. Higher values imply better force regulation.  
\textit{High [\%]} represents the fraction exceeding the optimal range, reflecting potential risks of object damage or actuator overload.  
Finally, \textit{MAE} measures deviation from the nominal optimal torque and thus serves as an indicator of overall grasping stability.  
Desired outcomes are characterized by an increase in \textit{Optimal [\%]} and decreases in \textit{Low [\%]}, \textit{High [\%]}, and \textit{MAE}.

\subsubsection{Subjective Metrics}

To assess perceived usability and workload, two standardized questionnaires were administered after each method:  
\begin{itemize}
  \item System Usability Scale (SUS)~\cite{sus}  
  \item NASA Task Load Index (NASA--TLX)~\cite{nasa-tlx}  
\end{itemize}
Additionally, participants provided open-ended feedback regarding perceived operability, confidence, and workload.  
All subjective and quantitative data were analyzed using paired Wilcoxon signed-rank tests, with significance thresholds set at $p<0.05$ and $p<0.01$.

\begin{figure}[t]
    \centering
    \includegraphics[width=0.9\linewidth]{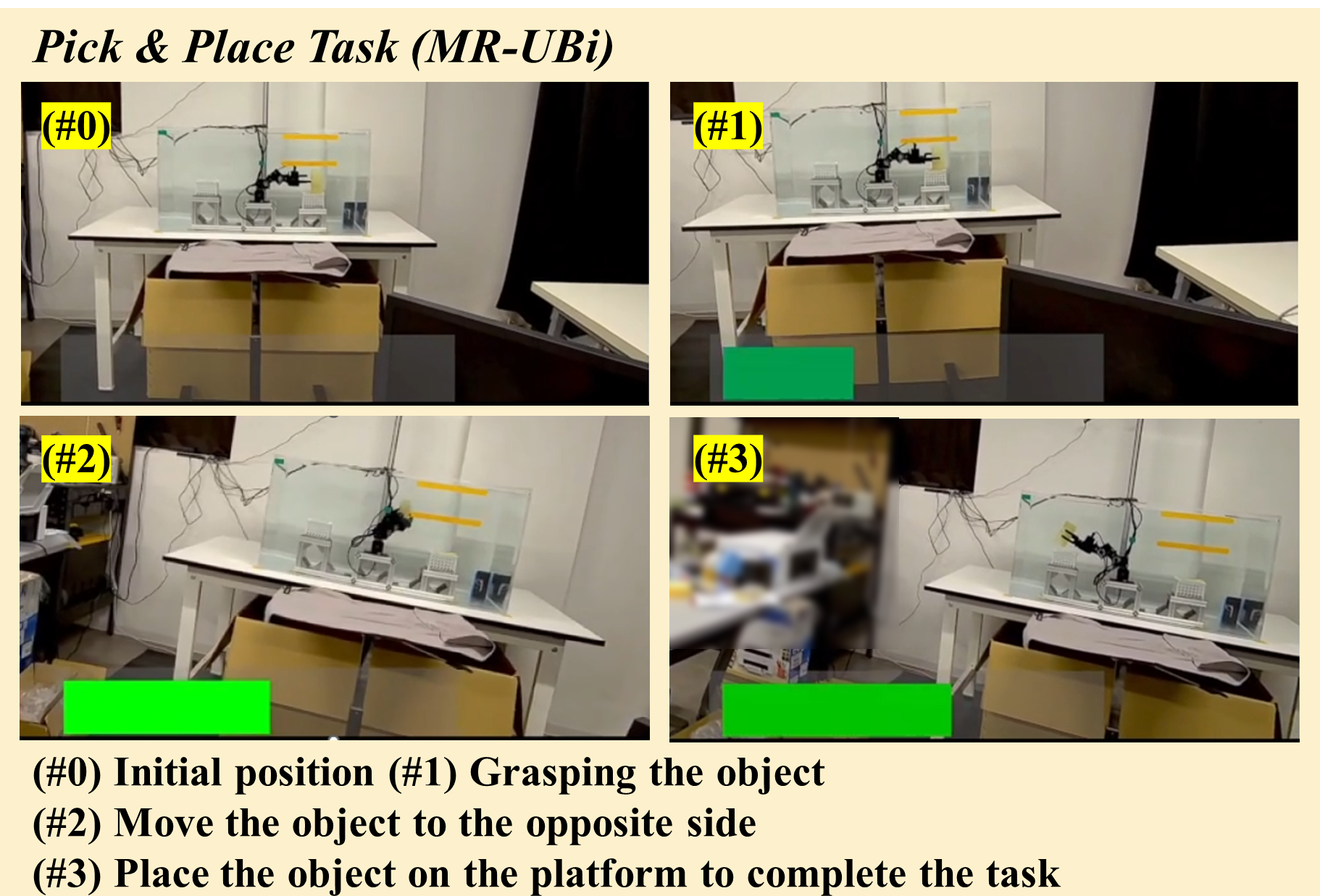}
    \caption{Pick\&Place Task using MR-UBi}
    \label{fig:MOVE_with_indica}
\end{figure}

\subsection{Experimental Results}
All tasks were successfully completed in both methods, while only the MR-UBi system’s operation examples are presented in Figs.~\ref{fig:LIFT_with_indica}–\ref{fig:MOVE_with_indica}.
\subsubsection{Quantitative Evaluation}
\begin{table}[t]
\centering
\caption{Results of Gripper Torque (J4).}
\label{tab:grip_summary}
\renewcommand{\arraystretch}{1.05}
\setlength{\tabcolsep}{4pt} 
\small
\begin{adjustbox}{width=\columnwidth}
\begin{tabular}{llccccc}
\toprule
\textbf{Task} & \textbf{Obj.} & \textbf{Method} & \textbf{Low [\%]} & \textbf{Opt [\%]} & \textbf{High [\%]} & \textbf{MAE [Nm]} \\
\midrule
\multirow{4}{*}{Lift}
  & \multirow{2}{*}{Block}  & Baseline & 39.5 & 30.0 & 30.5 & 0.200 \\
  &                         & MR-UBi   & \textbf{28.1} & \textbf{69.8} & \textbf{2.1} & \textbf{0.087} \\
  & \multirow{2}{*}{Sponge} & Baseline & 45.6 & 25.5 & 28.9 & 0.178 \\
  &                         & MR-UBi   & \textbf{28.6} & \textbf{69.5} & \textbf{1.9} & \textbf{0.088} \\
\midrule
\multirow{4}{*}{Pick\&Place}
  & \multirow{2}{*}{Block}  & Baseline & 37.6 & 29.7 & 32.8 & 0.191 \\
  &                         & MR-UBi   & \textbf{23.5} & \textbf{72.0} & \textbf{4.5} & \textbf{0.085} \\
  & \multirow{2}{*}{Sponge} & Baseline & 41.0 & 32.2 & 26.8 & 0.148 \\
  &                         & MR-UBi   & \textbf{23.1} & \textbf{75.5} & \textbf{1.4} & \textbf{0.077} \\
\bottomrule
\end{tabular}
\end{adjustbox}
\vspace{-2mm}
\end{table}
\begin{figure}[t]
    \centering
    \includegraphics[width=0.9\linewidth]{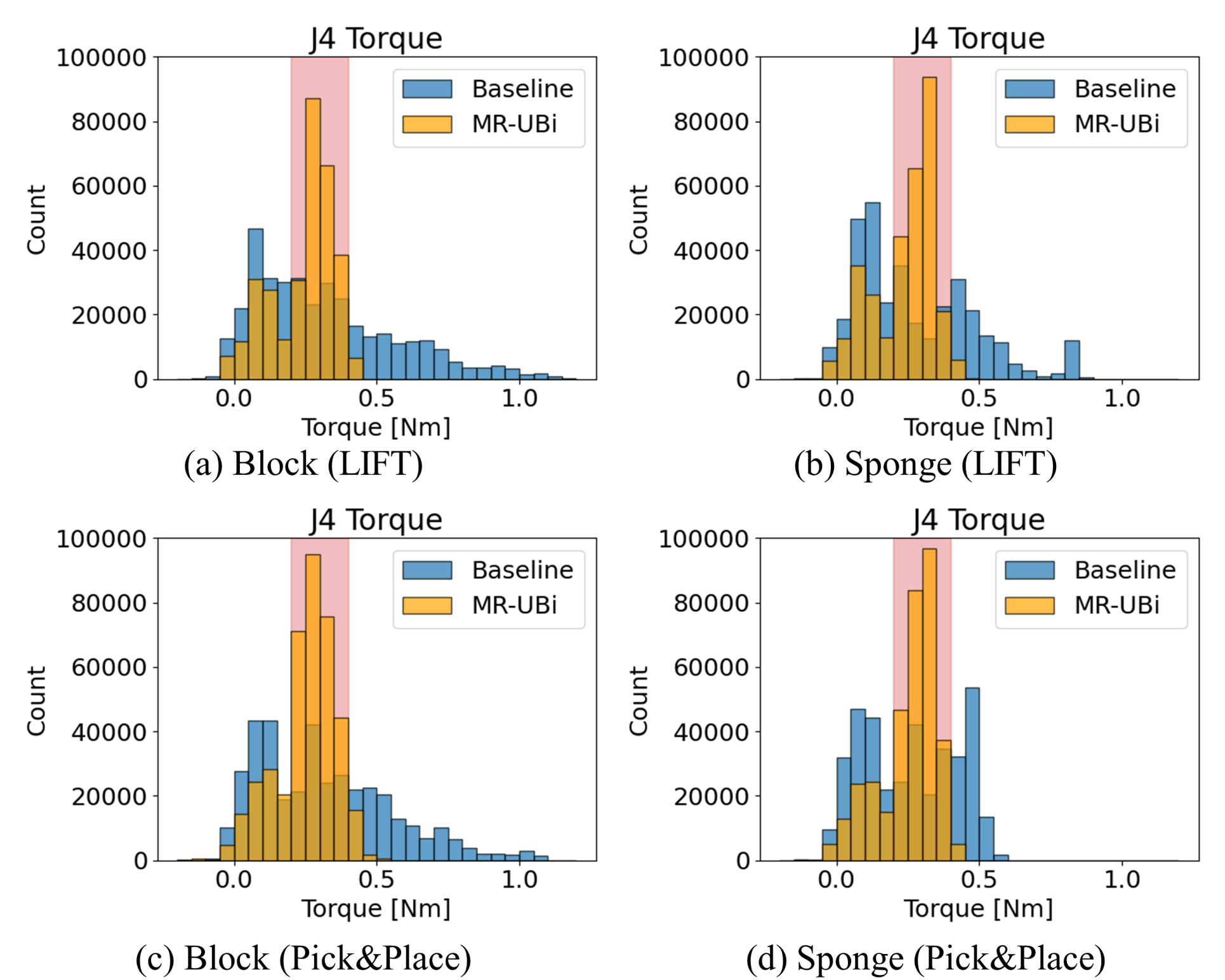}
    \caption{Comparison of Gripper Torque}
    \label{fig:joint3_grip_summary_all}
\end{figure}

Fig.~\ref{fig:joint3_grip_summary_all} shows torque distributions for each condition, object, and task; representative statistics are summarized in Table~\ref{tab:grip_summary}. 
Here, \textit{Optimal} denotes the proportion of samples within the target band, \textit{Low}/\textit{High} the proportions below/above the band, and \textit{MAE} the mean absolute error from the optimal torque.

\paragraph{Lift Task}
As illustrated in Fig.~\ref{fig:joint3_grip_summary_all} and Table~\ref{tab:grip_summary}, the \textit{Baseline} condition exhibits widely dispersed torques outside the optimal band. 
For the rigid block, we observed \textit{Optimal}~30.0\%, \textit{Low}~39.5\%, \textit{High}~30.5\%, and \textit{MAE}~0.200\,Nm. 
A similar trend appears for the compliant sponge: \textit{Optimal}~25.5\%, \textit{Low}~45.6\%, \textit{High}~28.9\%, and \textit{MAE}~0.178\,Nm, indicating a tendency toward both under- and over-application of force.

In contrast, \textit{MR-UBi} exhibits torque spread markedly contracts. 
For the block, \textit{Optimal} increases to 69.8\%, \textit{High} is suppressed to 2.1\%, and \textit{MAE} drops to 0.087\,Nm. 
For the sponge, \textit{Optimal} reaches 69.5\%, \textit{High} falls to 1.9\%, and \textit{MAE} is 0.088\,Nm, showing a pronounced concentration within the 0.20–0.40\,Nm band. 
These results indicate that the RTI curbs under/over-grasping and helps operators rapidly identify and maintain the appropriate torque range.

\paragraph{Pick\&Place Task}
In \textit{Baseline}, the block condition yields \textit{Optimal}~29.7\%, \textit{Low}~37.6\%, \textit{High}~32.8\%, and \textit{MAE}~0.191\,Nm, 
with frequent overshoot at motion onset/termination. 
For the sponge, \textit{Optimal}~32.2\%, \textit{Low}~41.0\%, \textit{High}~26.8\%, and \textit{MAE}~0.148\,Nm indicate substantial time outside the optimal band.

Under \textit{MR-UBi}, the block condition achieves \textit{Optimal}~72.0\%, \textit{High}~4.5\%, and \textit{MAE}~0.085\,Nm; 
the sponge condition reaches \textit{Optimal}~75.5\%, \textit{Low}~23.1\%, \textit{High}~1.4\%, and \textit{MAE}~0.077\,Nm, 
representing the most stable regulation across all settings. 
The reduction of excessive torque with the compliant object is especially notable, highlighting the RTI’s effectiveness in suppressing over-force events.

Taken together, these patterns hold across both objects and tasks, indicating robust gains in stability and accuracy with MR-UBi.
\begin{table}[t]
\centering
\caption{SUS and NASA--TLX Scores}
\label{tab:sus_nasa_combined}
\small
\renewcommand{\arraystretch}{1.05}
\setlength{\tabcolsep}{4.5pt}
\begin{adjustbox}{width=\columnwidth}
\begin{tabular}{lcccc}
\toprule
\textbf{Metric} & \textbf{Condition} & \textbf{Median} & \textbf{IQR} & \textbf{p-value} \\
\midrule
SUS & Baseline & 66.25 & 28.12 & 0.0006\,($p<0.01$) \\
    & MR-UBi   & \textbf{83.75} & \textbf{10.62} &  \\
\addlinespace[2pt]
NASA--TLX & Baseline & 49.50 & 27.92 & 0.0155\,($p<0.05$) \\
           & MR-UBi   & \textbf{41.17} & \textbf{20.25} &  \\
\bottomrule
\end{tabular}
\end{adjustbox}
\vspace{3pt}
\end{table}

\begin{figure}[t]
    \centering
    \includegraphics[width=\linewidth]{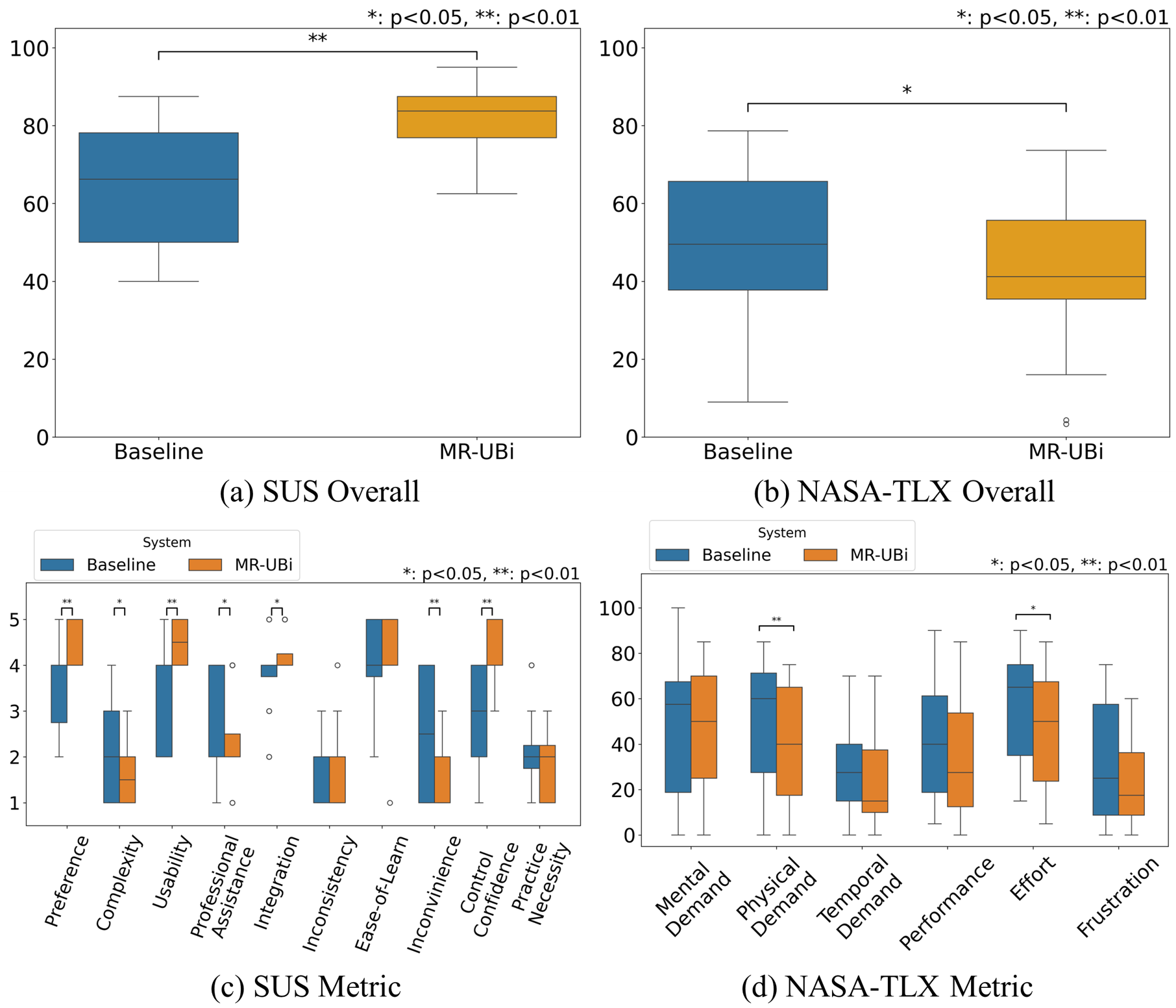}
    \caption{Results of SUS and NASA-TLX}
    \label{fig:sus_nasa_boxplots}
\end{figure}
\subsubsection{Subjective Evaluation (SUS and NASA--TLX)}
Fig.~\ref{fig:sus_nasa_boxplots} and Table~\ref{tab:sus_nasa_combined} summarize the SUS and NASA--TLX results.

\paragraph{System Usability Scale (SUS)}
Under \textit{Baseline}, the median overall score is 66.25 (IQR 28.12). 
With \textit{MR-UBi}, the median increases to 83.75 (IQR 10.62), a significant improvement over \textit{Baseline} ($p=0.000640$). 
Item-wise analyses show significant gains in \textit{Usability}, \textit{Control Confidence}, and \textit{Inconvenience} ($p<0.01$), indicating that the RTI boosts confidence while reducing hassle and stress. 
No significant differences are observed for \textit{Ease-of-Learn} and \textit{Practice Necessity}, suggesting. 
Overall, \textit{MR-UBi} significantly improves perceived usability.

\paragraph{NASA--TLX}
As shown in Table~\ref{tab:sus_nasa_combined}, the overall NASA--TLX score under \textit{MR-UBi} (41.17) is significantly lower than \textit{Baseline} (49.50) ($p<0.05$), 
indicating reduced perceived workload. 
Participants reported lower physical effort and effort-related burden, consistent with the RTI’s facilitation of torque tuning. 
Although \textit{Frustration} did not reach statistical significance, it showed a decreasing trend, suggesting potential mitigation of mental strain. 
No significant differences were found in \textit{Mental Demand} or \textit{Temporal Demand}, implying that the RTI does not increase cognitive or time pressure.

MR-UBi improved perceived operability and confidence (SUS) while reducing perceived workload (NASA--TLX), in agreement with the objective gains in torque regulation.

\subsection{Summary}
The experimental results confirmed both objectives. 
MR-UBi improved grasping-torque control accuracy by increasing the time within the optimal torque range and reducing torque error. 
It also enhanced perceived usability and reduced workload compared with the Baseline. 
Overall, MR-UBi enabled more stable, accurate, and user-friendly underwater teleoperation.

\section{Conclusion}
This paper presented MR-UBi, a novel mixed reality-based underwater robot arm teleoperation system with reaction torque indicator via bilateral control.  
By encoding reaction torque through the RTI, the MR-UBi enables intuitive force regulation even in underwater robot arm teleoperation.  
User studies with sixteen participants performing two manipulation tasks and objects demonstrated that MR-UBi significantly increased time within the optimal torque range and reduced both under- and over-grasping, while SUS and NASA--TLX results confirmed improved usability, confidence, and reduced workload.  
These findings verify that visual–haptic integration via MR-UBi facilitates stable and intuitive underwater robot arm teleoperation.
Experimental results validate the overall effectiveness of MR-UBi in real-world underwater tasks.

\bibliographystyle{IEEEtran}

\vfill

\end{document}